\documentclass[a4paper,twoside]{article}

\usepackage{epsfig}
\usepackage{subcaption}
\usepackage{calc}
\usepackage{amssymb}
\usepackage{amstext}
\usepackage{amsmath}
\usepackage{amsthm}
\usepackage{multicol}
\usepackage{pslatex}
\usepackage{apalike}
\usepackage{algorithm2e}
\usepackage[bottom]{footmisc}
\usepackage{SCITEPRESS}     

\begin{document}

\title{AccidentGPT: Large Multi-Modal Foundation Model for Traffic Accident Analysis}


\author{\authorname{Kebin Wu, Wenbin Li and Xiaofei Xiao}
\affiliation{Technology Innovation Institute, Abu Dhabi, United Arab Emirates}
\email{\{kebin.wu, wenbin.li, xiaofei.xiao\}@tii.ae}
}


\abstract{Traffic accident analysis is pivotal for enhancing public safety and developing road regulations. Traditional approaches, although widely used, are often constrained by manual analysis processes, subjective decisions, uni-modal outputs, as well as privacy issues related to sensitive data. This paper introduces the idea of AccidentGPT, a foundation model of traffic accident analysis, which incorporates multi-modal input data to automatically reconstruct the accident process video with dynamics details, and furthermore provide multi-task analysis with multi-modal outputs. The design of the AccidentGPT is empowered with a multi-modality prompt with feedback for task-oriented adaptability, a hybrid training schema to leverage labelled and unlabelled data, and  a edge-cloud split configuration for data privacy. 
To fully realize the functionalities of this model,
we proposes several research opportunities. This paper serves as the stepping stone to fill the gaps in traditional approaches of traffic accident analysis and attract the research community’s attention for automatic, objective, and privacy-preserving traffic accident analysis.}

\onecolumn \maketitle \normalsize \setcounter{footnote}{0} \vfill

\section{\uppercase{Introduction}}
\label{sec:introduction}

The rapid and accurate traffic accident analysis is critical in enhancing public safety and shaping effective road regulations. The tasks of the traffic accident analysis, varying from accident process reconstruction, responsibility attribution to traffic management and emergency response, are multifaceted and complex. Conventional approaches~\cite{Mohammed2019}, relying on eyewitness testimonies, official police documentation, and footage from surveillance cameras (if any), have been the core of the accident analysis for decades. However, these approaches are constrained by intensive manual labor nature, susceptibility to subjective biases, restricted uni-modal outputs, and the privacy concerns emerging from the handling of sensitive data~\cite{Al-ani2023}.

The advent of machine learning techniques have begun to boost the field of the traffic accident analysis, presenting enhanced precision and insights. Models are built by learning vast datasets including video footage, sensor data, and textual reports to achieve specific tasks such as accident detection~\cite{ALI2021105973}, accident prediction~\cite{CHAND20215135}, cause identification~\cite{NajafiMoghaddamGilani2021}. Focusing on the accident process reconstruction, from numerical modelling to software simulations of the collisions ~\cite{Duma_2022} are applied to determine sliced elements (e.g., pre-collision speed, traveled distance, trajectory) during the accident process. Nevertheless, these works are often uni-modal providing useful but fragmented information, while lacking the capacity to integrate and interpret diverse data sources cohesively to reconstruct all details (e.g., process video, vehicles' dynamics) of the accident and automate the post-accident management such as injury assessment, emergency response, report generation, and insurance claim. Furthermore, these traffic applications have been limited in their adaptability, often requiring extensive customization for each specific use case. 

As a step further, the recent emergence of the large language models (LLMs) such as LLaMa2~\cite{touvron2023llama} and large multi-modal models (LMMs) such as GPT-4V~\cite{wu2023nextgpt} not only demonstrate the capability to handle multi-modal inputs and outputs, but also underscores a paradigm shift towards task-agnostic learning frameworks, which generate insights across a myriad of tasks without the necessity for task-specific training. The intrinsic versatility of these models is manifested in their capability to generalize learned knowledge and skills across complex multi-task output scenarios. Although most of LMMs focus on dealing with image and text inputs and outputs, recent work~\cite{zhang2023metatransformer}, ~\cite{wu2023nextgpt} shows the possibility to bridge extended list of modalities (e.g., image, text, video, audio, video, etc.) as inputs and produce corresponding outputs of multi-modalities as a response to the prompt. In the context of traffic accident analysis, the technical foundation of such LMM models and techniques brings forward the possibility to build a foundation model to take into account multi-modal inputs and generate outputs for a multiplicity of traffic accident analysis tasks.  

However, while the incorporation of multi-modality in traffic analysis presents a promising frontier, it also brings to light significant challenges that have yet to be fully addressed specific to the field: 
\begin{itemize}
    \item Quality and Integrity of Data from Various Sources: In traffic analysis, data can come from a variety of sources, including dashcams, traffic cameras, eyewitness reports, vehicle sensors, and more. The quality and integrity of this data can vary greatly, impacting the accuracy and reliability of the analysis. The quality and integrity of the data from various sources are to be ensured for desired model performance; 
    \item Complexities with Seamless Interpreting and Reasoning: The complexities associated with seamless interpreting and reasoning from diverse traffic accident data and modalities are substantial; 
    \item Model Training and Task-Specific Outputs with Multi-Modal Inputs: The model training and the alignment of task-specific outputs with multi-modal inputs are challenging, which often require intricate customization and tuning; 
    \item Ethical and Privacy Concerns: Ethical and privacy concerns, especially related to the handling and processing of sensitive and personal data, have also been inadequately addressed. 
\end{itemize}

In this work, we propose the idea of AccidentGPT - a foundation model to transform the domain of traffic accident analysis by integrating multi-modal inputs not only to automatically reconstruct accident scenario details but also delivers comprehensive multi-task analysis with a variety of output modalities. The idea extends the existing LLM and LMM solutions with a multi-modality prompt coupled with a feedback mechanism for adaptive task optimization, a hybrid training schema leveraging both labelled and unlabelled data for enhanced model generalization and performance, and a edge-cloud split configuration for data privacy. This paper seeks to tackles the gaps in conventional solutions and unveil the potential of an automated, fast responding, objective, and privacy-preserving traffic accident analysis solution. 

The rest of the paper is organized as follows: section 2 discusses the gaps inherent in existing traffic accident analysis approaches. Section 3 outlines our idea of AccidentGPT, highlighting its multi-modal inputs and outputs and multi-task features. Section 4 presents the corresponding research opportunities. Section 5 concludes the paper.

\section{\uppercase{Gaps in Current Traffic Accident Analysis}}

The traditional approaches and contemporary machine learning techniques, while contributory, present several gaps and challenges that limit their applicability. These gaps highlight the urgent need for a systematic approach to traffic accident analysis.

\subsection{Data Integration and Analysis}

\textbf{Manual Efforts:} The traditional approaches~\cite{Mohammed2019} involve substantial manual efforts in post-accident data collection, processing, and analysis. This labor-intensive process is prone to bias of human judgment and thus can lead to inconsistencies and errors impacting the reliability of the analysis. Furthermore, the manual process is time-consuming and the lag in analysis undermines the immediacy of response, impacting emergency services, traffic management, and subsequent investigative processes. Automate the process with timeliness and systematic analysis is one of the key challenge to tackle. 

\textbf{Privacy Concerns:} Machine learning based approaches~\cite{NajafiMoghaddamGilani2021} integrates sensitive data sources (e.g., dashcam footage and bystander videos) and raise corresponding privacy and ethical concerns~\cite{8735700}. These challenges have constrained the scope and depth of accident analysis, leaving a wealth of potentially insightful data untapped. Ensuring the privacy of sensitive data directly improves the effectiveness of the traffic accident analysis.

\subsection{Model Modality and Generalization}

\textbf{Model Specialization:} Current machine learning models in the filed of traffic accident analysis are often specialized and task-specific~\cite{CHAND20215135}. These models excel in their designated tasks but face challenges when exposed to scenarios or data that are different from their training environments. The generalization capability of these models is limited, and this specialization hinders their adaptability and flexibility, reducing their applicability to a diverse range of accident scenarios and conditions. There exists a significant gap in developing models endowed with task-agnostic learning mechanisms that can seamlessly adapt and perform across a variety of tasks and conditions without the need for retraining or extensive customization.

\textbf{Uni-Modal Analysis:} Automatic traffic accident analysis on specific tasks~\cite{ALI2021105973} predominantly relies on uni-modal data sources, such as textual reports or image evidence. These uni-modal approaches lack the capacity to provide a holistic view of accident scenarios, often missing out on crucial contextual and dynamic information that multi-modal data can offer. The lack of versatility to adapt to different data types and analysis requirements leads to a fragmented and compartmentalized understanding of accident scenarios. There is a pressing need for models that can assimilate diverse data sources, understand the intricate interplay of dynamic factors, and provide a comprehensive analysis.

\textbf{Output Limitations:} The outputs of existing models~\cite{Duma_2022} for traffic accident analysis is typically limited in solo modality (e.g., responsibility, text report) as well. The uni-modality restricts the detailed insights that stakeholders, including investigators, traffic planners, and victims, can extract from the outputs. Furthermore, the lack of interoperability between different analysis systems and technologies can hinder comprehensiveness and intuitiveness of accident analysis across machine learning models. Models are expected to produce multi-modal outputs (e.g., visual representation, numerical dynamics, text reports and news) especially in a multi-task scenario in order to meet diverse stakeholders' requirements (e.g., responsibility attribution, video reconstruction) for a traffic accident analysis system. 

In the light of these gaps and challenges, this paper introduces AccidentGPT as a multi-modal foundation model capable of automatically interpreting a diverse range of data modalities and delivering comprehensive, multi-faceted outputs on multiple traffic accident analysis tasks.

\section{\uppercase{AccidentGPT Overview}}
\label{sec: general idea}

\begin{figure*}[t]
  \centering
   \includegraphics[width=1.0\linewidth]{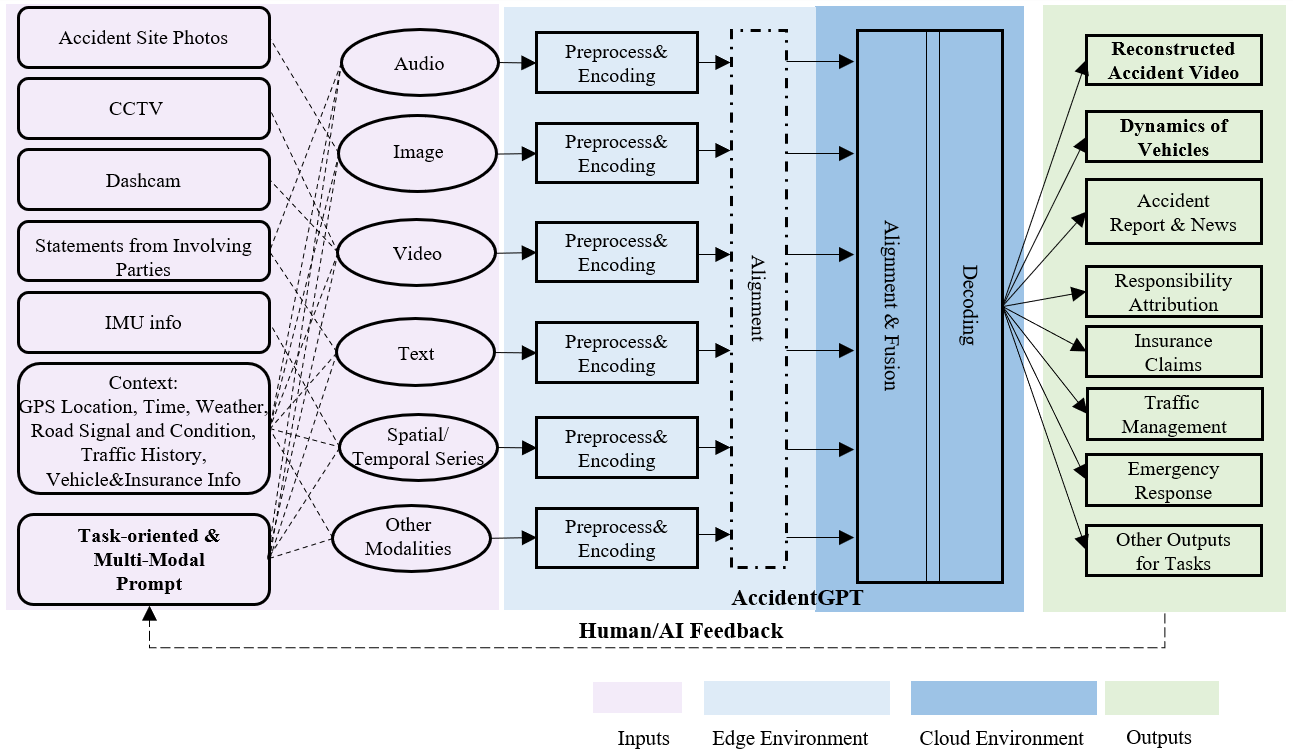}
   \caption{ AccidentGPT Overview.}
   \label{fig:framework}
\end{figure*}

The general idea of the AccidentGPT is depicted in Figure \ref{fig:framework}, and the model core follows a preprocessing\&encoding, alignment\&fusion, and decoding process. To revolutionize the field of traffic accident analysis, the joint of use of data from diverse sources is critical to provide robust and insightful analyses. The model inputs can include a) pre- and post-accident site photos, b) CCTV camera recordings, c) dashcam footage, d) statements about the accident process from the involving parties (e.g., drivers, witness) , e) information from Inertial Measurement Units (IMUs) of the movement dynamics during the accident, f) the contextual data containing the accident's related GPS location, time, road signal and condition (e.g., wet, dry, icy),  historical traffic data of the traffic sites, and the details related to vehicles and their insurances, and g) most importantly, the task-oriented prompt to instruct AccidentGPT on the desired analysis and outputs. AccidentGPT does not expect the comprehensive set of the input data in every accident analysis, but dynamically adapt to the available data for analysis with partial inputs as similar to the work ~\cite{moon2023anymal}. The statements, context and the prompt can be described in a multi-modal fashion (e.g., speech, text, image, etc.), leveraging both textual and non-textual data for a more holistic interpretation.

The model inputs encompass a variety of modalities, including audio, image, video, text, spatial and/or temporal tabular data, and other modalities such as remote sensing spectrum. Each input modality is subject to modality-specific preprocessing steps and encoders (e.g., CLAP~\cite{10095969} for audio, DinoV2~\cite{oquab2023dinov2} for image, AnyMAL-Video~\cite{moon2023anymal} for video, IMU2CLIP~\cite{moon2022imu2clip} for spatial/temporal series). 
During the model inference, the preprocessing and encoding process is to be carried out on the users' edge devices for the sake of privacy, and the decoding process is to be performed by the AccidentGPT model on cloud server. The edge machine learning techniques~\cite{LiW2023} can be applied to the encoders for computational efficiency and performance; in the case that specific encoders remain significantly demanding in computational resources after model compression, split learning~\cite{vepakomma2018split} can be leveraged by executing only the initial layers of the encoder on the edge devices, while the remaining layers can be offloaded and ran in the cloud environment. 

The alignment among different modalities~\cite{girdhar2023imagebind} harmonizes different data sources and ensures diverse modalities are properly integrated and correlated for accurate and cohesive analysis. In this edge-cloud split configuration, the alignment can be flexibly executed in the edge devices and/or the cloud environment. This provides adaptability based on the specific requirements of tasks, the computational resources available at the edge, and the desired response time. For simpler alignments on edge, the involving parties can quickly access the data and indicate the temporal and spatial properties of each data item (e.g., pre-accident, in-accident, post-accident). Conversely, for precise alignments involving multiple modalities and the complete input data, the cloud environment can be leveraged with superior computational capabilities to encapsulate the intricate cross-modal interactions among individual components spanning various modalities via representation fusion, coordination and fission~\cite{liang2023foundations}.

After preprocessing, encoding and alignment, the data are fed into the AccidentGPT for modality-specific decoding to automatically generate outputs corresponding to multiple tasks. AccidentGPT targets the following outputs of traffic accident analysis: 

\begin{itemize}
    \item \textbf{Reconstructed Video}: this output creates a visual 2D or 3D representation of the complete accident process. The resulting representation offers a temporal and spatially accurate depiction, providing investigators with a sequential understanding of the events leading up to, during, and after the accident.
    \item \textbf{Dynamics of Vehicles}: The dynamics of vehicles are associated with each video frame containing the following information of the involved vehicles: coordinates, velocity, direction, actions of each vehicle involved (i.e., braking, acceleration, turning, no action) and the point of impact and the damage descriptions. 
    \item \textbf{Accident Report \& News}: The accident report servers as the official documentation and details the sequence of events, involved parties, identified causes, and potential preventive measures. Based on the report, an accident news is tailored for dissemination to news agencies for public awareness.
    \item \textbf{Responsibility Attribution}: This output methodically identifies and attributes responsibilities to involved parties. 
    \item \textbf{Insurance Claims}: This output automates the insurance claim assessments by providing a data-driven breakdown of the accident, which highlights damages, identifies potential policy violations, and offers estimations of repair costs based on the severity and nature of the damages. 
    \item \textbf{Traffic Management}: This output primarily focuses on the immediate and long-term implications of traffic flow and infrastructure. Post-accident, the output provides real-time recommendations on traffic rerouting, crowd control, and area isolation to ensure minimal disruption and prevent secondary accidents. In the long term, based on recurrent patterns, the output identifies weaknesses in current infrastructure and traffic regulations, and suggest interventions to ensure smoother and safer traffic flow in the future.
    \item \textbf{Emergency Response}: Emphasizing immediate post-accident actions, this output assists in determining the severity of injuries, potential hazards (e.g., fuel leaks), and the requirement of specialized resources such as medical teams, fire brigades, or specialized rescue units. Additionally, the output provides essential information to first responders, like the number of vehicles involved, hazards, and the access to the accident site. This ensures that the response is not only swift but also tailored to the specific needs of the incident, minimizing harm and damage.
   \item \textbf{Other Tasks}: The adaptability and expansiveness of the AccidentGPT's design based on multi-modality and multi-task modelling makes the model suited for additional task-specific outputs not covered in the primary list. Such flexibility ensures that the model remains relevant and scalable, accommodating evolving traffic safety needs and technology advancements.    
\end{itemize}

Furthermore, the model provides avenues for multi-modal prompt based on reinforcement learning with human feedback (RLHF)~\cite{NIPS2017_d5e2c0ad} or AI feedback (RLAIF)~\cite{bai2022constitutional}, ensuring a continuous loop of learning and refinement to improve the model performance with the task-oriented and multi-modal prompt. 




\section{\uppercase{Research Opportunities}} \label{sec:oppo}

\subsection{Opportunity 1: Multi-Modal Traffic Data Collection and Integration}\label{ssec:data}

Gathering and integrating a comprehensive dataset for traffic accident analysis is essential for pretraining. 

\textbf{Collection and Standardization}: Similar to the paradigm shift in the computer vision domain with the introduction of ImageNet~\cite{5206848}, the traffic accident analysis field expects a transformation through the establishment of a comprehensive and standardized multi-modal dataset. However, the complexity and multifaceted nature of traffic accidents, along with the discrepancies in data collection methods across regions, make this endeavor challenging. The collaboration and standardization of the data gathering to consolidate such data require synchronized efforts from various stakeholders, in order to ensure the analysis uniformity and solution scalability. Alternatively,  leveraging simulation software for autonomous driving such as ~\cite{Cognata} can produce standardized datasets with controlled variables, which can act as a base for model training across different scenarios and conditions.

\textbf{Data Preprocessing}: Real-world traffic data can be noisy due to various interference sources like weather conditions affecting sensors or low-quality traffic cameras, and thus necessitate extensive proprocessing efforts (e.g., cleaning, filtering). On the other hand, collecting high-quality supervised data can be expensive and, at times, unfeasible. Although semi-supervised learning approaches resort to leveraging unlabeled or weakly labeled data, they still demand specialized filtering procedures ~\cite{radenovic2023filtering}. In a multi-modal scenario, the challenges compound even more, as each modality has its own inherent noise and discrepancies. The integration of varied data streams necessitates not only modality-specific preprocessing but also meticulous alignment and synchronization. This is essential to guarantee that inputs from disparate sources accurately represent a singular event. Additionally, an inter-modality harmonization mechanism~\cite{WU2023106457} is crucial to ensure that the composite representation holistically encapsulates the phenomenon under study, with no single modality disproportionately influencing the analysis.


\subsection{Opportunity 2: Multi-Modal Model Structure and Core Components}\label{ssec:model design}

Although the general idea of the AccidentGPT follows the encoding, alignment, fusion, and decoding process to generate multi-task multi-modal outputs, no dominant design of model structure exists yet either in existing vision-language pretraining~\cite{liu2023visual,alayrac2022flamingo,zhu2023minigpt,zhu2023minigpt4} or the multi-modality works~\cite{zhang2023metatransformer,wu2023nextgpt}, and no singular structure has been conclusively demonstrated to significantly outperform others. In addition to the model structure, the large multi-modal and multi-task model for traffic accident analysis involves four fundamental components that require further research innovations: 

\textbf{Alignment}: Alignment deals with the synchronization of data from various modalities to ensure that they represent the same event or phenomenon. Although emerging works~\cite{girdhar2023imagebind} demonstrate promising results, the extent and dimensions where the traffic accident data are shared across modalities can lead to: a) non-uniformity across modality alignment (e.g., one-to-one, one-to-many, or not exist at all), and b) long-range dependencies that a particular element from one modality corresponds to an element in another modality that is temporally or spatially distant. Effective alignment methods are expected for temporal matching, spatial calibration and semantic bridging across modalities.

\textbf{Fusion}: Once aligned, the fusion component form a unified representation of the data from different modalities and learn representations that capture the interactions between individual elements spanning various modalities~\cite{Man_2023_CVPR}. Due to the fact that multi-modal data are heterogeneous in characteristics, distribution, carried information and relevance toward specific tasks, this component is intrinsically challenging. In the filed of traffic accident analysis, the fusion process becomes even more important due to the critical spatial-temporal relations of the sequencing, timing, and positioning of events and actions that lead up to, occur during, and follow an accident.  


\textbf{Decoding}: The decoding process produces human-understandable outputs that reflect cross-modal interactions and coherence. While certain modality-specific decoders (e.g., text) are mature and widely used, the AccidentGPT decoding components do not merely construct raw outputs from model's internal representation, but also involves summarization of contents, translation between modalities and creation of new contents (i.e., reconstruction of accident process video). Video generation, as a modality, poses multiple challenges, especially when aiming for high fidelity and temporally coherent sequences. This is yet one of the most challenging but popular research direction. Recent advances~\cite{xu20234k4d} offer potential solutions, but further research is essential to enhance the granularity, accuracy, and realism of generated video content, particularly in the nuanced domain of traffic accident analysis due to its dynamics complexity, physical consistency and multi-modal integration.



\subsection{Opportunity 3: Multi-Modal Reasoning}\label{ssec:reasoning}
During the AccidentGPT's entire process of traffic accident analysis, reasoning with the fused representation is the key capability of to reconstruct the accident sequence, derive critical insights, and formulate logical conclusions about the incident's dynamics. The reasoning dimension is vast and complex for AccidentGPT involving the culmination of a series of events and interactions among multiple entities. The reasoning function is expected to: a) determine and learn the relationships and interactions within the accident scene, b) understand the contribution of each multi-modal data within the reasoning sequence, c) extrapolate increasingly abstract ideas from the individual pieces of multi-modal evidence. Existing works show contradicting results~\cite{stechly2023gpt4,huang-chang-2023-towards} on how well multi-modal models can perform on reasoning tasks, and yet a further step on the reasoning leveraging external large-scale knowledge and components can yield significant advancements in accurate accident reconstruction and understanding. 

\subsection{Opportunity 4: Data Efficient Training Paradigm}\label{ssec:pretraining}

Data from different sources for traffic accident analysis can be categorized into three types: labeled, unlabeled data, and weakly-labeled noisy data (even after preprocessing). Since the data related to traffic accident is scarce in general, it is worthwhile to investigate how to maximize the utilization of (pseudo) supervision or priors in the multi-modal data.  

\begin{figure}[t]
  \centering
   \includegraphics[width=1\linewidth]{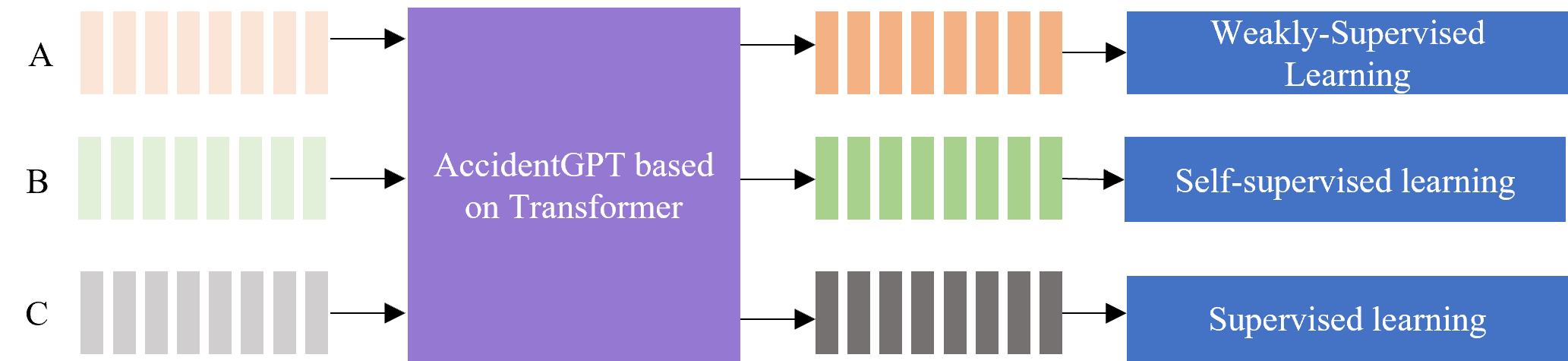}
   \caption{Framework for Pretraining.}
   \label{fig:training}
\end{figure}

One potential solution is to adopt a combined loss that enables supervised learning for labeled data \cite{dosovitskiy2020image}, self-supervised learning for unlabeled data \cite{he2022masked,chen2020simple} and weakly-supervised learning for weakly-labeled noisy data \cite{radford2021learning}. Without losing generalization, illustrative examples are shown in Figure \ref{fig:training}. In contrast to single strategy training, the hybrid training paradigm is another research opportunity allowing for the comprehensive exploitation of valuable and diverse data information, offering a flexible trade-off between the cost of data collection and the performance of the model. 

\subsection{Opportunity 5: Task-Oriented Multi-Modal Prompt with Feedback}\label{ssec:task-oriented}

The concept of "prompting" has demonstrated remarkable utility in LLMs~\cite{brown2020language} and LMMs~\cite{lyu2023macawllm}. By managing various tasks with task-specific descriptive prompts, attaching them to the input for downstream processing, and then jointly feeding them into a pre-trained, frozen foundational model, this approach offers a unified solution for diverse tasks. However, the full potential of prompting within the realm of large multi-modal models has yet to be fully explored. One of the paramount challenges lies in the vast complexity and diversity of multi-modal data. Unlike textual data where prompts can be relatively straightforward, defining an ideal prompt in multi-modal scenarios becomes intricate. And the alignment of task objectives with the modality specifics make the process non-trivial, as the misinterpretations or biases can have significant real-world consequences in the traffic accident analysis, ensuring the accuracy, interpretability, and contextual relevance of multi-modal prompts becomes absolutely critical.

Further, the process of feedback plays a vital role in shaping the effectiveness of the prompt. While RLHF~\cite{openai2023gpt4} can provide nuanced insights and guide the model towards desired outcomes, relying solely on it can be costly and time-consuming. On the other hand, recent work on RLAIF~\cite{bai2022constitutional} demonstrates that AI systems can potentially self-regulate, refine, and provide feedback only with the help of human oversight in terms of a list of rules or principles. This presents an intriguing paradigm where multi-modal prompts can be self-optimized and critiqued by a balance between human and AI feedback. The potential evolution of a feedback-driven prompting mechanism could pave the way for more granular and context-aware prompts, thereby enhancing the model's efficacy and responsiveness.

\subsection{Opportunity 6: Validation Methods and Reliability Metrics}\label{ssec:validation}
The evolution of multi-modal models in traffic accident analysis opens new avenues for research, particularly in the development of sophisticated validation techniques. Future studies should focus on creating methodologies that can accurately assess and ensure the reliability of outputs from complex systems like AccidentGPT. Another critical area of research is the formulation of robust metrics tailored to multi-modal, multi-task models in high-stake scenarios. These metrics would serve as benchmarks for evaluating the trustworthiness of the model's interpretations, which is paramount in traffic accident analysis.

\section{Conclusion}

In this paper, we have introduced AccidentGPT, an innovative foundation model tailored for the intricate domain of traffic accident analysis, leveraging multi-modal data streams and a multi-tasking paradigm. AccidentGPT synthesizes these varied data streams and processes them seamlessly through a unified analytical framework, thereby enabling comprehensive and insightful outputs that span multiple modalities and tasks. The potential of this approach represents a significant paradigm shift, promising to revolutionize the methodologies and tools available for traffic accident analysis.

Our work marks a first step towards an automatic, systematic and privacy preserving traffic accident analysis solution. Research efforts are required to refine these opportunities, fully realize their potential, and rigorously evaluate their performance in real-world scenarios. Future work will focus on exploring the related research opportunities and enhancing the effectiveness of the proposed approach.

\bibliographystyle{apalike}
{\small
\bibliography{example.bib}}

\end{document}